 \documentclass[pmlr,twocolumn,10pt]{jmlr} 

\usepackage{booktabs}
\usepackage{bm}
\usepackage{enumitem}
\usepackage{multicol}
\usepackage{subfiles}
\usepackage{amsmath}
\usepackage{amssymb}
\usepackage{relsize}
\usepackage{microtype}
\usepackage[font=scriptsize]{caption}
\usepackage{algpseudocode}
\usepackage{algorithm}
\algnewcommand\algorithmicforeach{\textbf{for each}}
\algdef{S}[FOR]{ForEach}[1]{\algorithmicforeach\ #1\ \algorithmicdo}

\usepackage{eqparbox}
\newdimen{\algindent}
\algnewcommand\LeftComment[2]{%
\hspace{#1\algindent}$\triangleright$ \eqparbox{COMMENT}{#2} \hfill %
}
\algnewcommand\LeftCommentNoTriangle[2]{%
\hspace{#1\algindent} \eqparbox{COMMENT}{#2} \hfill %
}
\algnewcommand\LeftCommentNoIntent[1]{%
$\triangleright$ \eqparbox{COMMENT}{#1} \hfill %
}

\newcommand{\StateOrange}[1]{\algrenewcommand{\alglinenumber}[1]{\footnotesize\textcolor{orange}{##1}:}\State #1}
\definecolor{darkpastelgreen}{rgb}{0.01, 0.75, 0.24}
\newcommand{\StateGreen}[1]{\algrenewcommand{\alglinenumber}[1]{\footnotesize\textcolor{darkpastelgreen}{##1}:}\State #1}

\definecolor{blue-violet}{rgb}{0.54, 0.17, 0.89}
\newcommand{\StateBlue}[1]{\algrenewcommand{\alglinenumber}[1]{\footnotesize\textcolor{blue-violet}{##1}:}\State #1}

\newcommand{\StateBlack}[1]{\algrenewcommand{\alglinenumber}[1]{\footnotesize\textcolor{black}{##1}:}\State #1}
\usepackage{verbatim}
\usepackage{tikz}
\usetikzlibrary{positioning,quotes,calc,fit,shapes,shadows,arrows}
\tikzset{block/.style={draw,very thick,text width=2cm,minimum height=4cm,align=center},
         line/.style={-latex}}
\tikzset{blockV/.style={draw,very thick,text width=2cm,minimum height=2cm, minimum width=4cm,align=center},
         line/.style={-latex}}
\tikzset{blockExt/.style={draw,very thick,minimum height=1cm, minimum width=1cm,align=center},
         line/.style={-latex}}
\usepgflibrary{arrows}
\usetikzlibrary{bayesnet}

\usepackage[load-configurations=version-1]{siunitx} 

\DeclareMathOperator*{\argminA}{argmin}

\theorembodyfont{\upshape}
\theoremheaderfont{\scshape}
\theorempostheader{:}
\theoremsep{\newline}

\jmlrvolume{LEAVE UNSET}
\jmlryear{2021}
\jmlrsubmitted{LEAVE UNSET}
\jmlrpublished{LEAVE UNSET}
\jmlrworkshop{Machine Learning for Health (ML4H) 2021} 

 \title{\hspace{1.7cm}Optimizing Binary Symptom Checkers \titlebreak ~via Approximate Message Passing}
\author{%
\Name{Mohamed Akrout}
\Email{mohamed@aip.ai}\\
\addr AIP Labs, Hungary
\AND
\Name{Faouzi Bellili, Amine Mezghani}
\Email{\{Faouzi.Bellili, Amine.Mezghani\}@umanitoba.ca}\\
\addr University of Manitoba, Canada
\AND
\Name{Hayet Amdouni} \Email{dr.amdounihayet@yahoo.fr}\\
\addr Self-employed Dermatologist, Tunisia
}
\begin{document}

\maketitle

\begin{abstract}
Symptom checkers have been widely adopted as an intelligent e-healthcare application during the ongoing pandemic crisis. Their performance have been limited by the fine-grained quality of the collected medical knowledge between symptom and diseases. While the binarization of the relationships between symptoms and diseases simplifies the data collection process, it also leads to non-convex optimization problems during the inference step. In this paper, we formulate the symptom checking problem as an underdertermined non-convex optimization problem, thereby justifying the use of the compressive sensing framework to solve it. We show that the generalized vector approximate message passing (G-VAMP) algorithm provides the best performance for binary symptom checkers.
\end{abstract}
\begin{keywords}
Symptom checkers, approximate message passing.
\end{keywords}

\vspace{-0.4cm}
\section{Introduction}
\begin{figure}[h]
\centering
\includegraphics[scale=0.3]{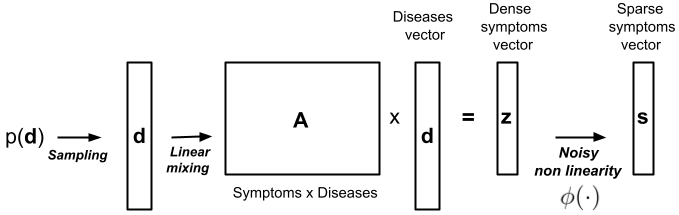}
\caption{The symptom checking model: a patient is represented by a disease vector $\mathbf{d}$ sampled from a predefined disease density $p(\mathbf{d})$, whose product with the matrix $\bm{A}$ containing the probabilistic relationships between symptoms and diseases leads to the dense symptom vector $\mathbf{z}$. The observed symptom vector $\mathbf{s}$ is a sparse version of $\mathbf{z}$ i.e. $\mathbf{s} = \phi(\mathbf{z + \mathbf{w}})$ where $\mathbf{w} \sim \mathcal{N}(\mathbf{w}; \bm{0}, \gamma_w^{-1}\,\mathbf{I})$.}
\label{fig:disease-diagnosis-process}
\vspace{-0.35cm}
\end{figure}

\vspace{-0.2cm}
With the growing shortage of doctors around the world, the geographical reach of the online symptom assessment tools, particularly symptom checkers, have provided alternative channels to better support patients at the very beginning of their diagnostic journey. The challenges faced by symptom checkers can be categorized into three classes:

\noindent\textbf{Medical knowledge data collection}: It is usually a tedious task to define the medical knowledge represented by the matrix $\mathbf{A}$ in Fig. \ref{fig:disease-diagnosis-process} since it requires a manual labor-intensive process to define the probabilistic relationships $p(s_i|d_j)$ between each symptom $s_i \in \mathcal{S}$ and disease $d_j \in \mathcal{D}$. While few studies (\cite{rotmensch2017learning}) explored the possibility to automate this process by learning high-quality knowledge bases associating diseases to symptoms directly from electronic medical records, the data-driven medical knowledge suffer from underestimating rare disease medical knowledge and requires an expert assessment.

\noindent\textbf{Sparse underdertermined inference}: The performance of most symptom checkers is limited by the number of observed symptoms which is usually a binary input, i.e., either the patient has the symptom or not. Additionally, the accuracy of the patient input is usually biased due to the medical vocabulary gap, incomplete information, and the inability to differentiate between highly correlated medical concepts. Moreover, the common design of symptom checkers usually require the number of conditions to be higher than the number of symptoms, i.e., $M < N$. This makes the problem in (\ref{eq:vamp-cost-function}) an underdertermined inference problem, which is precisely the typical assumption for recovering the sparse disease vector $\bm{d}$.

\noindent\textbf{Unbalanced datasets}: Symptom checkers based on machine learning algorithms such as supervised learning (\cite{pmlr-v56-Choi16}) and reinforcement learning (\cite{akrout2019improving}) are biased toward unbalanced datasets whose real-world relative frequencies of common vs. rare diseases are highly disproportional (\cite{fraser2017limitations}).

While early diagnosis is the most effective way to detect early diseases and reduce mortality (\cite{coleman2017early}), symptom checkers\footnote{See \href{https://symptoms.webmd.com/}{WebMD} and \href{https://symptomate.com/}{Symptomate}'s symptom checkers as examples.} still face three main challenges to accurately infer the correct disease $d_i^*$ among a set of $N$ supported diseases $\mathcal{D} = \{d_1,\, \dots,\, d_N\}$ given $K$ observed symptoms subset $\mathcal{S}_{\textrm{obs}}$ among a set of $M$ supported symptoms $\{s_1,\,\dots,\, s_M\}$, i.e., $\mathcal{S}_{\textrm{obs}} \subset \mathcal{S}$. That is to say:
\begin{equation}\label{eq:vamp-cost-function}
\small
     \widehat{\mathbf{d}}=\argminA_{\mathbf{d} \in \mathbb{R}^{N}} \frac{1}{2}\|\mathbf{s}-\phi(\mathbf{A}\, \mathbf{d})\|_{2}^{2}+f(\mathbf{d}).
\end{equation}
where $\phi(\cdot)$ is a non-linear function applied component-wise while $f(\mathbf{d})$ is a regularization function chosen to promote a desired structure in the unknown disease vector $\mathbf{d}$. For example, letting $f(\mathbf{d}) = \lambda\,\|\mathbf{d}\|_1$ with $\lambda > 0$ promotes sparsity in $\widehat{\mathbf{d}}$.

In this work, we propose a new methodology to jointly tackle the two challenges of symptom checkers by advocating the use of message passing algorithms for compressive sensing (\cite{donoho2009message}) to draw the following potential benefits:\vspace{0.2cm}

\noindent\textbf{Binary medical knowledge}: While collecting the probabilistic medical knowledge for each symptom-disease pair of the matrix $\mathbf{A}$ is a tedious task, collecting their binary version is more practical in terms of data collection speed and expert consensus convergence. In fact, most of the expert consensus issues lie in the magnitude of the relationship between any symptom-disease pair, which is automatically removed via the binarization.\vspace{0.2cm}

\noindent \textbf{Efficient algorithms for sparse inference}: The fact that the vector $\mathbf{d}$ is sparse in practice urges the need to use low-complexity compressive sensing algorithms such as GAMP (\cite{rangan2011generalized}) and VAMP (\cite{rangan2019vector}) which are well-suited for the under-complete problems in (\ref{eq:vamp-cost-function}) under non-convex sparsity-inducing penalty functions. The performance of symptom checkers based on sparse Bayesian inference are model-based algorithms which only rely on a predefined prior on the unknown disease vector $\mathbf{d}$.\vspace{-0.3cm}

\section{Dermatology case study}
To evaluate the performance of the proposed symptom checker based on message passing, we describe in this section our the data collection process for our dermatological study case. While many possible medical specialities were possible to explore, we decided to focus on dermatology due to the accessibility of the visual skin information, which will allow us to integrate the proposed symptom checker with visual classifiers in future works.\vspace{-0.3cm}
\subsection{Medical knowledge collection}
We collaborated with three practicing dermatologists to elicit the binary medical knowledge matrix $\mathbf{A}$ required for the symptom checker model in (\ref{eq:vamp-cost-function}). The participating doctors were encouraged to use the available medical literature and their experience in the process of creating the medical knowledge matrix. We explicitly asked them to provide the binary conditional probabilities $p(s_i|d_j)$, between each symptom $s_i \in \mathcal{S}$ and disease $d_j \in \mathcal{D}$ covering the following $M=27$ symptoms and $N=31$ diseases:

\noindent\textbf{Set of diseases} $\mathcal{D} = \{$psoriasis, seborrheic dermatitis, lichen, acne, atopic dermatitis, keloids, cheilitis, condylomata, candidiasis, dermatophyties, stasis dermatitis, dysidrosis, erysipelas, bedsores, folliculitis, hidradenitis suppurativa, cutaneous leishmaniasis, lupus erythematosus, melanoma, noevus, rosacea, toxidermia, ulcer venous, urticaria, varicella, herpes, zoster, sarcoidosis$\}$,

\noindent \textbf{Set of symptoms} $\mathcal{S} = \{$redness, dander, vesicles, bubbles, pigmentation, swelling, pustule, macule, plate, nodule, papule, crusts, hypochromia, atrophy, fever, pain, pruritus, oozing, hyperkeratosis, cracks, ulceration, ulcer, edema, induration, necrosis, infiltration, telangiectasias$\}$.


\begin{figure*}[ht!]
\centering
\begin{tikzpicture}[thick,scale=0.54, every node/.style={transform shape}]
  \node[block, fill=orange!15] (p_u) {\Large $p_\mathsf{\mathbf{d}}(\mathbf{d})$};
  \node[block, fill=blue!15, right= 5cm of p_u] (z_uv) {\Large $\mathbf{\mathsf{\mathbf{{z}}}=\mathbf{A}\,\mathsf{\mathbf{d}}}$};
  \node[block, fill=green!15, right= 5cm of z_uv,minimum width=3.5cm] (phi_z) {
  \mbox{\Large \hspace{-0.3cm}$\mathsf{\mathbf{{{s}}}} =\phi(\mathsf{\mathbf{{{z}}}} +\mathsf{\mathbf{{{w}}}})$}};
  \node[blockExt,right=of p_u, xshift=0.4cm, yshift=-1.5cm] (ext_pu_to_z_uv) {$\mathrm{\textbf{ext}}$};
  \node[blockExt,left=of z_uv, xshift=-0.4cm, yshift=1.5cm] (ext_z_uv_to_p_u) {$\mathrm{\textbf{ext}}$};
  \node[blockExt,right=of z_uv, xshift=0.4cm, yshift=-1.5cm] (ext_z_uv_to_y) {$\mathrm{\textbf{ext}}$};
  \node[blockExt,left=of phi_z, xshift=-0.4cm, yshift=1.5cm] (ext_y_to_z_uv) {$\mathrm{\textbf{ext}}$};
  \draw [-latex,very thick] ([yshift=-4.25em]p_u.east) -- 
  node [midway,below=0em,align=center ] { $\mathbf{\widehat{d}}^+_{\mathsf{p}}$}
  node [midway,below=1.6em,align=center ] {$\gamma_{\bm{\mathsf{d}}_{\mathsf{p}}}^+$}
  (ext_pu_to_z_uv.west);
  \draw [-latex,very thick] (ext_pu_to_z_uv) --
  node [midway,below=0em,align=center ] { $\mathbf{\widehat{d}}^+_{\mathsf{e}}$}
  node [midway,below=1.6em,align=center ] {$\gamma_{\bm{\mathsf{d}}_{\mathsf{e}}}^+$}
  ([yshift=-4.25em]z_uv.west)
  node [pos=0.25](ext_between_pu_z_uv){};
  \draw [-latex,very thick] ([yshift=-4.25em]z_uv.east) --
  node [midway,below=0em,align=center ] { $\mathbf{\widehat{z}}^+_{\mathsf{p}}$}
  node [midway,below=1.6em,align=center ] {$\gamma_{\bm{\mathsf{z}}_{\mathsf{p}}}^+$}
  (ext_z_uv_to_y.west);
  \draw [-latex,very thick] (ext_z_uv_to_y) --
  node [midway,below=0em,align=center ] { $\mathbf{\widehat{z}}^+_{\mathsf{e}}$}
  node [midway,below=1.7em,align=center ] {$\gamma_{\bm{\mathsf{z}}_{\mathsf{e}}}^+$}
  ([yshift=-4.25em]phi_z.west)
  node [pos=0.25](ext_between_z_uv_y){};
  \draw [-latex,very thick] ([yshift=4.25em]phi_z.west) --
  node [midway,above=0em,align=center ] { $\mathbf{\widehat{z}}^-_{\mathsf{p}}$}
  node [midway,above=1.6em,align=center ] {$\gamma_{\bm{\mathsf{z}}_{\mathsf{p}}}^-$}
  (ext_y_to_z_uv.east);
  \draw [-latex,very thick] (ext_y_to_z_uv) --
  node [midway,above=0em,align=center ] { $\mathbf{\widehat{z}}^-_{\mathsf{e}}$}
  node [midway,above=1.7em,align=center ] {$\gamma_{\bm{\mathsf{z}}_{\mathsf{e}}}^-$}
  ([yshift=4.25em]z_uv.east)
  node [pos=0.25](ext_between_y_z_uv){};
  \draw [-latex,very thick] ([yshift=4.25em]z_uv.west) --
  node [midway,above=0em,align=center ] { $\mathbf{\widehat{d}}^-_{\mathsf{p}}$}
  node [midway,above=1.7em,align=center ] {$\gamma_{\bm{\mathsf{d}}_{\mathsf{p}}}^-$}
  (ext_z_uv_to_p_u.east);
  \draw [-latex,very thick] (ext_z_uv_to_p_u) --
  node [midway,above=0em,align=center ] { $\mathbf{\widehat{d}}^-_{\mathsf{e}}$}
  node [midway,above=1.7em,align=center ] {$\gamma_{\bm{\mathsf{d}}_{\mathsf{e}}}^-$}
  ([yshift=4.25em]p_u.east)
  node [pos=0.25](ext_between_z_uv_pu){};
  \draw [-latex,very thick] (ext_between_pu_z_uv.center) --
  (ext_z_uv_to_p_u.south);
  \draw [-latex,very thick] (ext_between_z_uv_y.center) --
  (ext_y_to_z_uv.south);
  \draw [-latex,very thick] (ext_between_y_z_uv.center) --
  (ext_z_uv_to_y.north);
  \draw [-latex,very thick] (ext_between_z_uv_pu.center) --
  (ext_pu_to_z_uv.north);
\end{tikzpicture}
\caption{Block diagram of the generalized VAMP algorithm with its three modules: the prior modules $p_\mathsf{\mathbf{{{d}}}}(\bm{d})$, the linear minimum mean square estimator (LMMSE) module $\mathbf{\mathsf{\mathbf{{{z}}}}=\mathbf{A}\,\mathsf{\textbf{{{d}}}}}$ and the non linear block $\mathsf{\mathbf{{s}}} =\phi(\mathsf{\mathbf{{{z}}}} +\mathsf{\mathbf{{{w}}}})$. Each module uses the turbo principle (widely known in iterative decoding literature) to pass the extrinsic messages to its adjacent module.}
\label{fig:block-diagram}
\end{figure*}
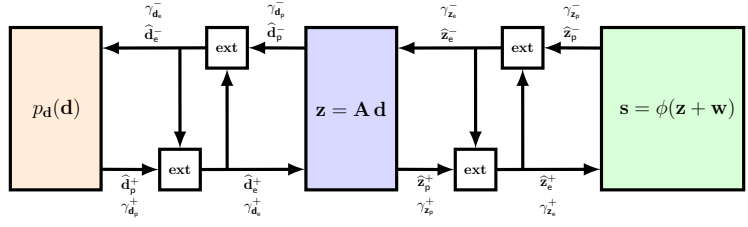
\vspace{-0.3cm}
\subsection{Evaluation of vignettes collection}
Assessing the performance of the proposed symptom checker requires the collection of a real-world test set containing the medical consultation information, a.k.a., a ``vignette'', of patients represented by:\vspace{-0.25cm}
\begin{itemize}[leftmargin=*]
    \item The list of symptoms, $\mathcal{S}_{\textrm{obs}}$, of the patient described by the medical consultation, which is different from the list of symptoms reported by the dermatologist. This choice guarantees that the symptom vector $\mathbf{s}$ takes into account the partial observability of the set of manifested symptoms of the patient due to his/her limited medical knowledge.\vspace{-0.25cm}
    \item The ground truth disease, $d_i^*$, of the patient confirmed by the dermatologist after one or multiple medical consultations. Therefore, the target disease vector $\mathbf{d}$ contains one non-zero element equal to 1 in the $i$th position.
\end{itemize}
At the end of the process, under the patients' consent, we collected 200 vignettes covering the 31 diseases, i.e., $\sim$ 7 vignettes per disease.\vspace{-0.3cm}

\section{Approximate message passing optimization}
\vspace{-0.2cm}
Approximate message passing (AMP)-based computational techniques have gained a lot of attention since their introduction within the compressed sensing framework (\cite{donoho2009message}) to solve optimization problems involving a linear mixing of the vector to be optimized, as in (\ref{eq:vamp-cost-function}). In particular, vector AMP (VAMP) (\cite{rangan2019vector}) and generalized VAMP (G-VAMP) (\cite{schniter2016vector}) algorithms exhibit faster convergence for a broad (non i.i.d.) class of sensing matrices that are right orthogonally-invariant, unlike the generalized AMP (GAMP) algorithm (\cite{rangan2011generalized}).
\begin{figure*}[ht!]
\scriptsize
\begin{multicols}{2}
  \begin{equation}
    \label{eq:denoising-function-d}
      \bm{g}_{\bm{\mathsf{d}}}\left(\mathbf{\widehat{d}}, \gamma_{\bm{\mathsf{d}}}\right) = \frac{ \mathlarger{\int} \boldsymbol{d}\,\,p_{\bm{\mathsf{d}}}\left(\boldsymbol{d}\right)\,\mathcal{N}\big(\boldsymbol{d}; \mathbf{\widehat{d}}, \gamma_{\bm{\mathsf{d}}}^{-1}\bm{I}\big) \,\textrm{d}\boldsymbol{d}}{\mathlarger{\int} p_{\bm{\mathsf{d}}}\left(\boldsymbol{d}\right)\,\mathcal{N}\big(\boldsymbol{d}; \mathbf{\widehat{d}}, \gamma_{\bm{\mathsf{d}}}^{-1}\bm{I}\big)\, \textrm{d}\boldsymbol{d}}
  \end{equation}\break
  \begin{equation}
    \label{eq:denoising-function-z-non-linear}
    \bm{g}_{\bm{\mathsf{z}}}\left(\mathbf{\widehat{z}}, \gamma_{\bm{\mathsf{z}}},\mathbf{s}\right) = \frac{ \mathlarger{\int} \boldsymbol{z}\,\,p_{\bm{\mathsf{s}}|\bm{\mathsf{z}}}\left(\boldsymbol{s}|\boldsymbol{z}\right)\,\mathcal{N}\big(\boldsymbol{z}; \mathbf{\widehat{z}}, \gamma_{\bm{\mathsf{z}}}^{-1}\bm{I}\big) \,\textrm{d}\boldsymbol{z}}{\mathlarger{\int} p_{\bm{\mathsf{s}}|\bm{\mathsf{z}}}\left(\boldsymbol{s}|\boldsymbol{z}\right)\,\mathcal{N}\big(\boldsymbol{z}; \mathbf{\widehat{z}}, \gamma_{\bm{\mathsf{z}}}^{-1}\bm{I}\big)\, \textrm{d}\boldsymbol{z}}
  \end{equation}
\end{multicols}
\vspace{-0.5cm}
\begin{equation}
\label{eq:denoising-function-z}
\hspace{0cm}\bm{f}_{\bm{\mathsf{r}}}\left(\mathbf{\widehat{z}}, \gamma_{\bm{\mathsf{z}}}, \mathbf{\widehat{d}}, \gamma_{\bm{\mathsf{d}}}\right) = \frac{ \mathlarger{\int}\mathlarger{\int} \boldsymbol{r}\,\,\mathcal{N}\big(\boldsymbol{d}; \mathbf{\widehat{d}}, \gamma_{\bm{\mathsf{d}}}^{-1}\bm{I}\big)\,\mathcal{N}\big(\boldsymbol{z}; \mathbf{\widehat{z}}, \gamma_{\bm{\mathsf{z}}}^{-1}\bm{I}\big)\,\mathcal{N}\big(\boldsymbol{z}|\boldsymbol{d}; \mathbf{\bm{A}\,\boldsymbol{d}}, \gamma_{\bm{\mathsf{z}}}^{-1}\bm{I}\big) \,\textrm{d}\boldsymbol{z}\,\textrm{d}\boldsymbol{d}}{\mathlarger{\int}\mathlarger{\int} \mathcal{N}\big(\boldsymbol{d}; \mathbf{\widehat{d}}, \gamma_{\bm{\mathsf{d}}}^{-1}\bm{I}\big)\,\mathcal{N}\big(\boldsymbol{z}; \mathbf{\widehat{z}}, \gamma_{\bm{\mathsf{z}}}^{-1}\bm{I}\big)\,\mathcal{N}\big(\boldsymbol{z}|\boldsymbol{d}; \mathbf{\bm{A}\,\boldsymbol{d}}, \gamma_{\bm{\mathsf{z}}}^{-1}\bm{I}\big) \,\textrm{d}\boldsymbol{z}\,\textrm{d}\boldsymbol{d}}, ~\text{with}~ \bm{\mathsf{r}} = \{\bm{\mathsf{d}}, \bm{\mathsf{z}}\}.\vspace{-0.7cm}
\end{equation}
\end{figure*}
One could also consider the use of alternating direction method of multipliers (ADMM) (\cite{boyd2011distributed}), which is a penalty-based iterative algorithm for solving constrained optimization problems. However, it is known that VAMP converges faster than ADMM because of its automatic tuning of the augmented Lagrangian parameter $\rho$ in ADMM (\cite{manoel2018approximate}). In fact, one of the variable splitting strategies employed by ADMM, known as Douglas-Rachford splitting (\cite{douglas1956numerical}), corresponds to VAMP by letting $\rho = \gamma_{\bm{\mathsf{d}}_{\mathsf{p}}}^+/2$ where $\gamma_{\bm{\mathsf{d}}_{\mathsf{p}}}^+$ is the precision estimate of $\mathbf{d}$.\vspace{0.1cm}

\noindent \textbf{VAMP algorithm for symptom checkers:}
For better illustration, the block diagram of the generalized VAMP (G-VAMP) algorithm and its use toward solving the symptom checkers optimization problem in (\ref{eq:vamp-cost-function}) is depicted in Fig. \ref{fig:block-diagram}. There, we show its different constituent blocks, namely the different denoisers as they interact with the so-called LMMSE module. All the exchanged extrinsic information is performed using expectation propagation (a.k.a., turbo) principle which approximates the posterior messages by Gaussian distributions. The algorithmic steps of the G-VAMP are summarized in 
Algorithm \ref{alg:vamp-algorithm} where each line number is colored similarly to the corresponding module in Fig. \ref{fig:block-diagram}. The subscript $t$ stands for the iteration index and subscripts $\mathsf{p}$ and $\mathsf{e}$ are used to distinguish “posterior” and “extrinsic” variables, respectively.
\begin{algorithm}[h!]
   \caption{\small{Generalized VAMP for symptom checkers}}
   \label{alg:vamp-algorithm}
\begin{algorithmic}[1]
\fontsize{7.2}{7.2}\selectfont
\Statex $\mathbf{Input:}$ observed symptom vector $\boldsymbol{s} \in\mathbb{R}^{M}$; medical knowledge matrix $\mathbf{A}$; precision tolerance $\xi=10^{-6}$; number of iterations $T_\textrm{max}$; noise precision $\gamma_w$. 
\State $\mathbf{Initialize}$ 
\Statex $t\gets 1$
\Statex\LeftCommentNoIntent{initialize posterior and extrinsic means and precisions}
\Statex $\boldsymbol{\widehat{d}}^{-}_{\mathsf{e},0}$, $\gamma_{\bm{\mathsf{d}}_{\mathsf{e}},0}^-$,
$\boldsymbol{\widehat{z}}^{-}_{\mathsf{e},0}$, $\gamma_{\bm{\mathsf{z}}_{\mathsf{e}},0}^-$\vspace{0.1cm}
\State \textbf{Repeat}\vspace{0.1cm}
\Statex \par \LeftCommentNoIntent{denoising $\bm{d}$}
\StateOrange $\widehat{\bm{d}}_{\mathsf{p}, t}^+=\bm{g}_{\bm{\mathsf{d}}}\left(\mathbf{\widehat{d}}^-_{\mathsf{e},t-1}, \gamma_{\bm{\mathsf{d}}_{\mathsf{e}, t-1}}^-\right)$\label{algo:denoising1-d}
\StateOrange $\alpha_{\bm{\mathsf{d}}_\mathsf{p}, t}^+=\left\langle \bm{g}_{\bm{\mathsf{d}}}^{\prime}\left(\mathbf{\widehat{d}}^-_{\mathsf{e},t-1}, \gamma_{\bm{\mathsf{d}}_{\mathsf{e}, t-1}}^-\right)\right\rangle$
\StateOrange $\gamma_{\bm{\mathsf{d}}_{\mathsf{p}}, t}^+=\gamma_{\bm{\mathsf{d}}_{\mathsf{e}},t-1}^- / \alpha_{\bm{\mathsf{d}}_\mathsf{p}, t}^+$, \quad$\gamma_{\bm{\mathsf{d}}_{\mathsf{e}},t}^+=\gamma_{\bm{\mathsf{d}}_{\mathsf{p}}}^+-\gamma_{\bm{\mathsf{d}}_{\mathsf{e}, t-1}}^-$
\StateOrange $\mathbf{\widehat{d}}^+_{\mathsf{e}, t}=\left(\gamma_{\bm{\mathsf{d}}_{\mathsf{p},t}}^+\, \widehat{\bm{d}}_{\mathsf{p}, t}^+-\gamma_{\bm{\mathsf{d}}_{\mathsf{e}, t-1}}^-\, \mathbf{\widehat{d}}^-_{\mathsf{e},t-1}\right) / \gamma_{\bm{\mathsf{d}}_{\mathsf{e}},t}^+$
\StateBlue $\widehat{\bm{d}}_{\mathsf{p}, t}^-=\bm{f}_{\bm{\mathsf{d}}}\left(\mathbf{\widehat{z}}^-_{\mathsf{e},t-1}, \gamma_{\bm{\mathsf{z}}_{\mathsf{e}, t-1}}^-, \mathbf{\widehat{d}}^+_{\mathsf{e}, t}, \gamma_{\bm{\mathsf{d}}_{\mathsf{e}},t}^+\right)$
\StateBlue $\alpha_{\bm{\mathsf{d}}_\mathsf{p}, t}^-=\left\langle \bm{f}_{\bm{\mathsf{d}}}\left(\mathbf{\widehat{z}}^-_{\mathsf{e},t-1}, \gamma_{\bm{\mathsf{z}}_{\mathsf{e}, t-1}}^-, \mathbf{\widehat{d}}^+_{\mathsf{e}, t}, \gamma_{\bm{\mathsf{d}}_{\mathsf{e}},t}^+\right)\right\rangle$
\StateBlue $\gamma_{\bm{\mathsf{d}}_{\mathsf{p}}, t}^-=\gamma_{\bm{\mathsf{d}}_{\mathsf{e}},t}^+ / \alpha_{\bm{\mathsf{d}}_\mathsf{p}, t}^-$, \quad$\gamma_{\bm{\mathsf{d}}_{\mathsf{e}},t}^-=\gamma_{\bm{\mathsf{d}}_{\mathsf{p}}}^--\gamma_{\bm{\mathsf{d}}_{\mathsf{e}, t}}^+$
\StateBlue $\mathbf{\widehat{d}}^-_{\mathsf{e}, t}=\left(\gamma_{\bm{\mathsf{d}}_{\mathsf{p},t}}^-\, \widehat{\bm{d}}_{\mathsf{p}, t}^--\gamma_{\bm{\mathsf{d}}_{\mathsf{e}, t}}^+\, \mathbf{\widehat{d}}^+_{\mathsf{e},t}\right) / \gamma_{\bm{\mathsf{d}}_{\mathsf{e}},t}^-$
\Statex\LeftCommentNoIntent{denoising $\bm{z}$}
\StateBlue $\widehat{\bm{z}}_{\mathsf{p}, t}^+=\bm{f}_{\bm{\mathsf{z}}}\left(\mathbf{\widehat{z}}^-_{\mathsf{e},t-1}, \gamma_{\bm{\mathsf{z}}_{\mathsf{e}, t-1}}^-, \mathbf{\widehat{d}}^+_{\mathsf{e}, t}, \gamma_{\bm{\mathsf{d}}_{\mathsf{e}},t}^+\right)$ \label{algo:denoising-d}
\StateBlue $\alpha_{\bm{\mathsf{z}}_\mathsf{p}, t}^+=\left\langle \bm{f}_{\bm{\mathsf{z}}}^{\prime}\left(\mathbf{\widehat{z}}^-_{\mathsf{e},t-1}, \gamma_{\bm{\mathsf{z}}_{\mathsf{e}, t-1}}^-,\mathbf{\widehat{d}}^+_{\mathsf{e}, t}, \gamma_{\bm{\mathsf{d}}_{\mathsf{e}},t}^+\right)\right\rangle$
\StateBlue $\gamma_{\bm{\mathsf{z}}_{\mathsf{p}}, t}^+=\gamma_{\bm{\mathsf{z}}_{\mathsf{e}},t-1}^- / \alpha_{\bm{\mathsf{z}}_\mathsf{p}, t}^+$, \quad$\gamma_{\bm{\mathsf{z}}_{\mathsf{e}},t}^+=\gamma_{\bm{\mathsf{z}}_{\mathsf{p}}}^+-\gamma_{\bm{\mathsf{z}}_{\mathsf{e}, t-1}}^-$
\StateBlue $\mathbf{\widehat{z}}^+_{\mathsf{e}, t}=\left(\gamma_{\bm{\mathsf{z}}_{\mathsf{p},t}}^+\, \widehat{\bm{z}}_{\mathsf{p}, t}^+-\gamma_{\bm{\mathsf{z}}_{\mathsf{e}, t-1}}^-\, \mathbf{\widehat{z}}^-_{\mathsf{e},t-1}\right) / \gamma_{\bm{\mathsf{z}}_{\mathsf{e}},t}^+$\vspace{0.1cm}
\Statex\LeftCommentNoIntent{Estimation of  $\bm{z}$ from $\bm{s}$}\vspace{0.1cm}
\StateGreen $\widehat{\bm{z}}_{\mathsf{p}, t}^-=\bm{g}_{\bm{\mathsf{z}}}\left(\mathbf{\widehat{z}}^+_{\mathsf{e},t}, \gamma_{\bm{\mathsf{z}}_{\mathsf{e},t}}^+,\mathbf{s}\right)$
\StateGreen $\alpha_{\bm{\mathsf{z}}_\mathsf{p}, t}^-=\left\langle \bm{g}_{\bm{\mathsf{z}}}^{\prime}\left(\mathbf{\widehat{z}}^+_{\mathsf{e},t}, \gamma_{\bm{\mathsf{z}}_{\mathsf{e}, t}}^+,\mathbf{s}\right)\right\rangle$
\StateGreen $\gamma_{\bm{\mathsf{z}}_{\mathsf{p}}, t}^-=\gamma_{\bm{\mathsf{z}}_{\mathsf{e}},t}^+ / \alpha_{\bm{\mathsf{z}}_\mathsf{p}, t}^-$, \quad$\gamma_{\bm{\mathsf{z}}_{\mathsf{e}},t}^-=\gamma_{\bm{\mathsf{z}}_{\mathsf{p}}}^--\gamma_{\bm{\mathsf{z}}_{\mathsf{e}, t-1}}^+$
\StateGreen $\mathbf{\widehat{z}}^-_{\mathsf{e}, t}=\left(\gamma_{\bm{\mathsf{z}}_{\mathsf{p},t}}^-\, \widehat{\bm{z}}_{\mathsf{p}, t}^--\gamma_{\bm{\mathsf{z}}_{\mathsf{e}, t}}^+\, \mathbf{\widehat{z}}^+_{\mathsf{e},t}\right) / \gamma_{\bm{\mathsf{z}}_{\mathsf{e}},t}^-$
\StateBlack $t \gets t + 1$\vspace{0.1cm}
\State \textbf{Until}\,$\Big(\big|\!\big|\boldsymbol{\widehat{d}}_{ \mathsf{p},t+1}^{+}-\boldsymbol{\widehat{d}}_{ \mathsf{p},t}^{+}\big|\!\big|^2_{\textrm{F}}\Big)$ $\leq\xi\Big( \big|\!\big|\boldsymbol{\widehat{d}}_{ \mathsf{p},t}^{+}\big|\!\big|^2_{\textrm{F}}$\Big)~\text{or}~ \Big($t>T_\textrm{max}\Big)$\vspace{0.1cm}
\State \textbf{return} $\widehat{\bm{d}}_{\mathsf{p}, T_{\textrm{max}}+1}^+$
\end{algorithmic}
\end{algorithm}

\noindent \textbf{Denoising $\bm{d}$ and $\bm{z}$:} The denoising functions of $\mathbf{d}$ and $\mathbf{z}$ in lines \ref{algo:denoising1-d} and \ref{algo:denoising-d} are given in (\ref{eq:denoising-function-d}) and (\ref{eq:denoising-function-z}), respectively. The function $\bm{g}_{\bm{\mathsf{d}}}\left(\cdot, \cdot\right)$ in (\ref{eq:denoising-function-d}) acts as a “denoiser” of the additive white Gaussian noise-corrupted pseudo-measurement $\mathbf{\widehat{d}}^-_{\mathsf{e}} = \bm{d}^* + \mathbf{w}$ where $\mathbf{w}\sim\mathcal{N}(\bm{w};\bm{0},\bm{I}/\gamma_{\bm{\mathsf{d}}_{\mathsf{e}}}^{-})$, using the prior knowledge on the disease ground truth $\bm{d}^*$. The denoising bloc in purple in Fig. \ref{fig:block-diagram} denoises both $\mathbf{d}$ and $\mathbf{z}$ under the pseudo-prior $\mathcal{N}(\bm{d};\mathbf{\widehat{d}}^+_{\mathsf{e}},\bm{I}/\gamma_{\bm{\mathsf{d}}_{\mathsf{e}}}^+)$ and $\mathcal{N}(\bm{z};\mathbf{\widehat{z}}^-_{\mathsf{e}},\bm{I}/\gamma_{\bm{\mathsf{z}}_{\mathsf{e}}}^-)$, respectively, using the function $\bm{f}_{\bm{\mathsf{r}}}\left(\mathbf{\widehat{z}}, \gamma_{\bm{\mathsf{z}}}, \mathbf{\widehat{d}}, \gamma_{\bm{\mathsf{d}}}\right)$ given in (\ref{eq:denoising-function-z}) where $\bm{\mathsf{r}} = \{\bm{\mathsf{d}}, \bm{\mathsf{z}}\}$.\vspace{0.1cm}

\noindent \textbf{Estimating $\bm{z}$ from symptoms $\bm{s}$:}
The fact that the generative model of the observed symptom vector $s$ has the form $\mathsf{\mathbf{{s}}} =\phi(\mathsf{\mathbf{{{z}}}} +\mathsf{\mathbf{{{w}}}})$ allows the calculation of the posterior mean $\mathbf{\widehat{z}}^-_{\mathsf{p}}$ using the function $\bm{g}_{\bm{\mathsf{z}}}(\cdot)$ given in (\ref{eq:denoising-function-z-non-linear}). There, the density $p_{\bm{\mathsf{s}}|\bm{\mathsf{z}}}\left(\boldsymbol{s}|\boldsymbol{z}\right)$ represents the non-linear element-wise channel through which the dense symptom vector $\bm{z}$ is transformed in a sparse symptom vector $\bm{s}$. For binary symptom checkers, we let $\phi(x) = \textrm{sign}(x)$ which corresponds to the following denoising function:\vspace{-0.2cm}
\begin{equation}\label{eq:1bit-channel}
\bm{g}_{\bm{\mathsf{z}}}\left(\mathbf{\widehat{z}}, \gamma_{\bm{\mathsf{z}}},\mathbf{s}\right)=
\begin{cases}
\frac{v}{u}+\frac{u}{\gamma_{\bm{\mathsf{z}}}}\cdot \frac{\phi(v)}{\Phi(v)} &, \textrm{if} \quad s_i = 1 \\
\frac{v}{u}-\frac{u}{\gamma_{\bm{\mathsf{z}}}}\cdot \frac{\phi(v)}{1-\Phi(v)} &, \textrm{if} \quad s_i = -1
\end{cases}
\end{equation}
\vspace{-0.5cm}
\begin{subequations}
\begin{align*}
    \textrm{where}~~ &u =\frac{1}{\sqrt{\gamma_{\bm{\mathsf{z}}}^{-2}+\gamma_w^{-2}}}, \quad v =\mathbf{\widehat{z}}\cdot u\\
    &\phi(x) =\frac{1}{\sqrt{2 \pi}} e^{-\frac{1}{2} x^{2}}, \quad\Phi(x) =\int_{-\infty}^{x} \phi(t) \,\textrm{d}t
\end{align*}
\end{subequations}

\vspace{-0.6cm}
\section{Evaluation \& Results}
\vspace{-0.15cm}
We evaluate the NRMSE performance of G-VAMP (\cite{rangan2019vector}) on the $P = 200$ collected vignettes. We benchmark it with other optimization techniques, namely: $i)$ sparse sensing (\cite{rozenberg2018sparse}) which works by scanning all possible support bases for the disease vector $\bm{d}$ when it has either 1 or 2 non-zero elements, $ii)$ ADMM (\cite{boyd2011distributed}) for a range values of augmented Lagrangian penalty $\rho \in [0,5]$ with a step size of 0.1, and $iii)$ the standrad underdetermined least square (ULS) closed-form solution (\cite{datta2010numerical}).
We also evaluate the symptom checker using the top-K accuracy metric for $K \in \{1,2,3\}$, which is well known in computer vision classification scenarios to successfully assess the differential diagnosis cases. All G-VAMP experiments were run with a signal-to-noise ratio (SNR) of 25 dB.

Results are reported in Table~\ref{tab:NRMSE-topk}. We observe that G-VAMP outperforms the three baseline algorithms significantly. While ADMM exhibits a performance close to G-VAMP, this has been achieved by reporting its best performance among all 51 experiments by varying its parameter $\rho \in [0,5]$  with a step size of 0.1. Clearly, the auto-tuning feature of G-VAMP not only provides a better performance but also reduces the time complexity of symptom checkers which must provide a real-time result for the end user.\vspace{-0.1cm}

\begin{table}[ht]
\scriptsize
\centering
\caption{NRMSE and top-K metrics of the different algorithms on the binary symptom checker problem (\ref{eq:vamp-cost-function}).}
\vspace{-0.2cm}
\begin{tabular}[t]{ccccc}
\cmidrule[0.028cm]{2-5}
&\textbf{ULS}&\textbf{Sparse sensing}&\textbf{ADMM}&\textbf{G-VAMP}\\
\midrule
NRMSE&0.70&0.56&0.53&\textbf{0.44}\\ \midrule
TOP 1&0.39&-&0.78&\textbf{0.81}\\ \midrule
TOP 2&0.41&-&0.81&\textbf{0.83}\\ \midrule
TOP 3&0.43&-&0.84&\textbf{0.87}\\
\bottomrule
\end{tabular}
\label{tab:NRMSE-topk}
\end{table}

\vspace{-0.5cm}
\section{Future work \& Conclusion}
\vspace{-0.15cm}
To simplify the medical knowledge data collection, we have formulated binary symptom checkers as an optimization problem and advocated AMP algorithms to solve it by showing a significant improvement in terms of NRMSE and top-K accuracy performance. Many avenues for further investigation remain open. When the prior knowledge of the disease density on $\bm{d}$ is not available, it is possible to relax this requirement using data-driven methods to infer the prior from the available datasets. It is also possible to substitute the denoising function $\mathbf{g}_{\bm{\mathsf{d}}}(\cdot)$ with a separate optimization procedure via Plug-and-Play priors (\cite{venkatakrishnan2013plug}). One can also investigate the bilinear problem version of (\ref{eq:vamp-cost-function}) to jointly recover both the disease vector $\mathbf{d}$ and the medical knowledge matrix $\mathbf{A}$ by employing bilinear message passing algorithms such as BiG-VAMP (\cite{akrout2020bilinear}) and BiG-AMP (\cite{parker2014bilinear}). This will enable the comparison of the reconstructed matrix $\mathbf{A}$ with the one designed by medical experts.
\clearpage
\bibliography{jmlr-sample}
\end{document}